# Illumination Normalization via Merging Locally Enhanced Textures for Robust Face Recognition

Chaobing Zheng[1], Shiqian Wu[1,2], Wangming Xu[1,2*], Shoulie Xie[3]

[1] Wuhan University of Science and Technology, Wuhan, People's Republic of China
[2] Institute of Robotics and Intelligent Systems, Wuhan University of Science and Technology, Wuhan, People's Republic of China
[3] Institute for Infocomm Research, 1 Fusionopolis Way, 21-01 Connexis, Singapore
*xuwangming@wust.edu.cn

**Abstract:** In order to improve the accuracy of face recognition under varying illumination conditions, a local texture enhanced illumination normalization method based on fusion of differential filtering images (FDFI-LTEIN) is proposed to weaken the influence caused by illumination changes. Firstly, the dynamic range of the face image in dark or shadowed regions is expanded by logarithmic transformation. Then, the global contrast enhanced face image is convoled with difference of Gaussian filters and difference of bilateral filters, and the filtered images are weighted and merged using a coefficient selection rule based on the standard deviation (SD) of image, which can enhance image texture information while filtering out most noise. Finally, the local contrast equalization (LCE) is performed on the fused face image to reduce the influence caused by over or under saturated pixel values in highlight or dark regions. Experimental results on the Extended Yale B face database and CMU PIE face database demonstrate that the proposed method is more robust to illumination changes and achieve higher recognition accuracy when compared with other illumination normalization methods and a deep CNNs based illumination invariant face recognition method.

## 1. Introduction

Face recognition has been widely applied in many fields including public security, law enforcement and commerce, access control, information security, intelligent surveillance and so on. The reason is that it has the advantages of simplicity, quickness, no disturbance, and non-contact acquisition, when compared with fingerprint, iris, vein and other biometric recognition technologies. Currently face recognition products available on the market perform well only in well-controlled illumination conditions. Once lighting changes, the performance will be severely degraded [1,2].

Illumination variation is one of the most significant factors that reduce the performance of face recognition. Adini's empirical study shows that none of the image representations is sufficient by itself to overcome image variations because of a change in the direction of illumination [3]. Since then, a lot of research work have been made to implement illumination normalization for face recognition [1,2,4,5], and many of them have reported promising results on some public database. Shan et al. proposed the illumination normalization by localizing the holistic approaches in [6]. They processed image in local blocks, which can strengthen the local information and weakens the influence caused by side-lighting. With the increase of the block size, the results tend to become more and more smooth, thus the effect caused by side-lighting become smaller, but some useful information will also lose simultaneously Vishwakarma et al. used a fuzzy filter in DCT domain to normalize illumination in [7]. They found that the effect of illumination variations is in decreasing order over low-frequency discrete cosine transform (DCT) coefficients. By using the fuzzy filter, it can nullify the effect of illumination variations as well as to preserve the low-frequency details of a face image under high and unpredictable illumination variations. Kakadiaris et al. proposed a 3D-2D framework for illumination normalization and face recognition in [8]. They used 2D texture data and 3D shape to construct a new 3D face to normalize local illumination between probe and gallery textures. The new images are robust against illumination and pose variations.

Recently deep learning has been increasingly concerned in pattern recognition. Ramaiah et al. applied the convolution neural networks into the face recognition under different illumination conditions [10], and achieved good results. Li et al. used a convolutional neural network to estimate illumination map called Lighten-Net in [9], then reconstructed the image based on Retinex. This method is characterized by using a data-driven way to enhance weakly illuminated image, its performance is better than the hand-crafted methods. However deep learning technique generally requires a considerable amount of data and special hardware to train and deploy in practice, which is not suitable for embedded and low power devices.

Among these face image illumination normalization methods, most of them perform perfectly under well-controlled illumination conditions, but they are still deficient under less controlled illumination conditions. To solve this problem, a fusion of differential filtering images based local texture enhanced illumination normalization method (FDFI-LTEIN) is designed to weaken the influence caused by illumination variation in this paper. The main contributions of the paper are summarized as follows:

・Some existing illumination normalization methods are analyzed and divided into three categories.

・A local texture enhanced illumination normalization method based on fusion of differential filtering images (FDFI-LTEIN) is proposed and compared with some existing illumination normalization methods.

・Multi-scale local uniform binary patterns histograms (MSULBPH) is proposed. It is fully evaluated with PCA, LDA and LBPH through different illumination normalization methods on two public face databases.

The remainder of this paper is organized as follows: Section 2 briefly introduces different illumination preprocessing approaches for lighting-invariant face recognition. Section 3 describes the proposed method in detail. Experimental results are shown in Section 4. Some conclusions about this work are given in the last section.

## 2. Illumination Normalization Approaches

Compared to pose and facial expressions changes, illumination variation is more frequent, and it is more difficult to be controlled under unconstrained conditions. Related work shows that the intrapersonal differences in face appearances due to lighting variation can be much larger than interpersonal differences [11]. Thus illumination variations increase intra-class distance and reduce the correct rate of face recognition. Before giving our solution to this problem, we first review several related illumination normalization methods for face image in this section.

In recent years, numerous illumination normalization methods have been proposed [2]. They can be classified into three categories [6]: gray-level transformation, robust illumination information extraction, and reflectance model estimation, as shown in Table 1.

**Table 1** Illumination Normalization Approaches

| Column heading | Column heading two |
| --- | --- |
| Gray-level Transformation | HE[12] CLAHE[13] |
| Robust Illumination Information Extraction | GIC[14] |
|  | LoG[15]  DGD[15] |
| Reflection Model Estimation | SSR[16] ASR[17] LDCT[17] |
|  | TT[18] LTV[19 |

### 2.1. Gray-level Transformation

The principle of gray-level transformation is to process the pixel values of the image directly. Implementation is very simple and efficiency. Gray-level transformation can be further divided into linear and nonlinear. Usually nonlinear method is more effective than linear method to enhance the contrast of face images. It includes histogram equalization (HE)[12], adaptive histogram equalization (AHE)[22], contrast limited adaptive histogram equalization (CLAHE)[13], Gamma intensity correction (GIC)[14], logarithmic transformation (LT), index transformation. Linear method includes: Gray-level linear transformation, segmented linear transformation and so on. Gray-level transformation can stretch the face image contrast effectively, enhance the dynamic range of the image in dark or shadowed regions, and make the details of the face more visibility. But they cannot remove the side lighting effect essentially.

### 2.2. Robust Illumination Information Extraction

Nastar et al. pointed out that the changing light mainly affects the low spectrum of the image in [23]. Also the effect of illumination on the different frequency segments of the image is different. The principle of robust illumination information extraction is to extract the gray-level gradients or edges from face images and use them as a lighting-insensitive representation. Commonly used methods include: Directional gray-level derivative (DGD) [15], Laplace of Gaussian (LoG) [15], gradient faces normalization (GN) [23] and so on. Such algorithms are robust to lighting to some extent. But under the varying lighting conditions such as strong shadows and overexposure, facial features are blurry. Hence, the information of images is severely lost, and greatly degrades the face recognition performance. Therefore, they are not suitable for extracting the gray-level gradients or edges as a lighting-insensitive representation.

## 2.3. Reflectance Model Estimation

The principle of reflectance model estimation is to build a model and estimate the face reflectance field from a 2D face image [6]. One category of such methods is based on Retinex theory, which is one of the most classical algorithms[16], and took the lead in solving the problem of illumination change. It was put forward by Landis to explain the color perception property of the human vision system, which assumes the observed color image $I(x, y)$ can be decomposed into reflectance and illumination $I(x, y) = R(x, y)L(x, y)$, where $R(x, y)$ is the reflectance, and $L(x, y)$ denotes the illumination intensity at the point $(x, y)$. $L(x, y)$ is determined by the illumination source, $R(x, y)$ is determined by the characteristics of the surface of the objects. Therefore, the illumination normalization for face recognition can be achieved by estimating the illumination and then dividing the image, which can be given by:

$$r(x, y) = \log R(x, y) = \log \frac{I(x, y)}{L(x, y)} \tag{1}$$

$$L(x, y) = F(x, y) * I(x, y) \quad F(x, y) = \lambda e^{\frac{-(x^2+y^2)}{c^2}} \tag{2}$$

$$r(x, y) = \log I(x, y) - \log[F(x, y) * I(x, y)] \tag{3}$$

Although this method has achieved some progresses on illumination estimation under some complex lighting conditions, they still have some limitations. For example, strong shadows cast from a direct light source and the illuminations on the light shielding edges usually change quickly, which violate the assumption that the illumination slowly varies. Moreover, halo or shadow effects are often visible at large illumination discontinuities in $I$. To solve these problems, a series of improvement strategies have been proposed such as mutli-scale retinex (MSR) [25], adaptive single scale retinex (ASR) [18], homomorphism filtering (HOMO) [26], discrete cosine transformation (DCT) [19], Tan and Triggs (TT) [20] and etc.

Among these methods, TT performs better than others. Hence we briefly introduce TT method here. In TT, there are three main steps, e.g. gamma correction (GC), difference of Gaussian (DoG) [27] filtering and contrast equalization (CE). TT can enhance local texture information of facial image under varying lighting conditions to some extent. But it may produce halo under the condition of side-lighting as shown Fig 1, which has a negative effect on face recognition.

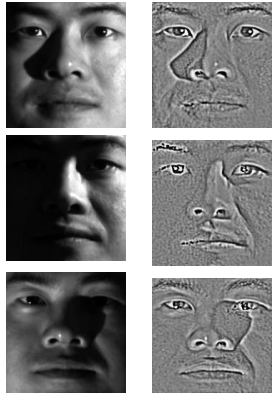

*Fig. 1.* *Original images are in the left row, Images processed by TT are in the second row*

## 3. The proposed Methods

It can be seen from Fig. 1 that the shadow is enhanced by TT. The reason is that both shadows and facial features would have the same gradient magnitudes. In the second row, it still suffers from halo artifacts under the condition of side-lighting. To solve this problem, a local texture enhanced illumination normalization based on fusion of differential filtering images (FDFI-LTEIN) is proposed in this paper，which can enhance local textures under different lighting conditions without enhanced interference noise. The proposed illumination normalization method is illustrated in Fig 2.

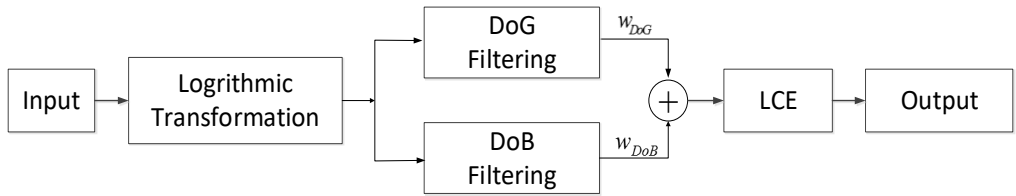

*Fig. 2. The stages of our illumination normalization*

### 3.1. Logarithmic Transformation

Logarithmic transformation is a nonlinear gray-level transformation that replaces gray-level $I$ with $\log_c(I+\varepsilon)$. Both logarithmic transformation and gamma transformation can enhance the local dynamic range of the image in dark or shadowed regions, as shown in Fig 3, but the dynamic range of $log_2(I+\varepsilon)$ is wider than $In(I+\varepsilon)$, $gamma(I,0.2)$ and $lg(I+\varepsilon)$. Hence it does not only stretch the image contrast, but also can enhance the details of the face image information. The result of the different gray-level transformation methods have been shown in Fig 4, we can see that $log_2(I+\varepsilon)$ and $In(I+\varepsilon)$ can make face images more balanced than $gamma(I,0.2)$. To make face image look more stable, we select $log_2(I+\varepsilon)$ instead of $In(I+\varepsilon)$. Where $\varepsilon$ is a small value to avoid logarithm of zero, so we replace gray-level $log_2(I)$ with $log_2(I+\varepsilon)$. Then, we can get the gray-level transformation image $I' = \log_2(I+\varepsilon)$.

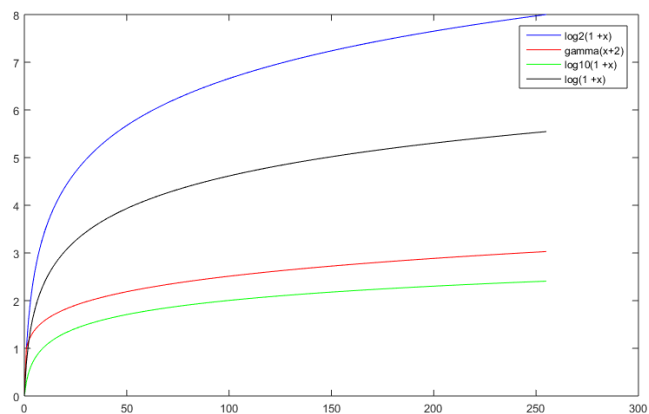

*Fig. 3.* *Different gray-level enhancement*

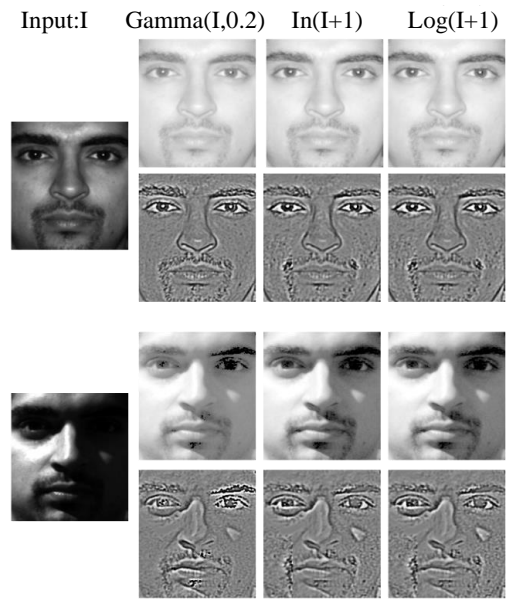

*Fig. 4.* *Face image processed by different gray-level enhancement*

### 3.2. DoG and DoB filtering

As a feature enhancement algorithm, DoG can be utilized to increase the visibility of edges and other details in a digital image. But it is not an edge-preserving smoothing filter for images, so that some details about the border are lost. But bilateral filter is a non-linear, edge-preserving and noise-reducing smoothing filter for images. Because of the characteristics of the edge-preserving filter, the filter assigns very low weights to the neighborhood pixels that have large differences between their luminance values and the value of the center pixel, thus the halo artifacts are significantly reduced. The intensity value at each pixel in an image is replaced by a weighted average of intensity values from nearby pixels. Moreover, the weights depend not only on euclidean distance of pixels, but also on the radiometric differences (e.g. range differences, such as color intensity, depth distance, etc.). This preserves sharp edges by systematically looping through each pixel and adjusting weights to the adjacent pixels accordingly.

Spatial closeness weight:

$$d(x,y,k,l) = \exp(\frac{(x-k)^2 + (y-l)^2}{2\sigma_d^2}) \tag{4}$$

Intensity difference weight:

$$r(x,y,k,l) = \exp(\frac{\left\|I^{'}(x,y) - I^{'}(k,l)\right\|^2}{2\sigma_r^2}) \tag{5}$$

Where $\sigma_d$ and $\sigma_r$ are smoothing parameters, $I^{'}(x,y)$ and $I^{'}(k,l)$ are the intensity of pixels $(x,y)$ and $(k,l)$, respectively. Then, the weight assigned for pixel $(x,y)$ to denoise the pixel $(k,l)$ is given by

$$w(x,y,k,l) = d(x,y,k,l)gr(x,y,k,l) = \exp(\frac{(x-k)^2 + (y-l)^2}{2\sigma_d^2} - \frac{\left\|I^{'}(x,y) - I^{'}(k,l)\right\|^2}{2\sigma_r^2}) \tag{6}$$

After calculating the weights, we can get the denosied image:

$$I_B(x,y) = I^{'}(x,y) * w(x,y,k,l) = \frac{\sum\sum_{(k,l)\in\Omega} I^{'}(k,l) \cdot w(x,y,k,l)}{\sum\sum_{(k,l)\in\Omega} w(x,y,k,l)} \tag{7}$$

where $\Omega$ is a local neighbourhood of the pixel (x, y), $R \times R$ is size of neighbourhood, points $(k, l)$ indicate the locations of pixels in the neighbourhood. Then we can get the difference of bilateral (DoB) [28] filtering by using two different $w$ as follows

$$I_{DoB} = I_{B1} - I_{B2} = I' * w_1 - I' * w_2 \tag{8}$$

DoB is also a is a non-linear, edge-preserving and noise-reducing smoothing filter, which can increase the visibility of edges and other details with no loss of edge and other useful information.

Compared with the DoG, DoB adds intensity difference weight. It can preserve the edge details of the facial image without producing halo. So much high-frequency information can be preserved by using DoB. But noise is often high-frequency, such as the boundary between shadow region and non-shadow region produced in uneven light condition. It will have a bad effect on face recognition. Gaussian filter can suppress high frequency signal and minimize boundary effects caused by side-lighting. DoG can work as a bandpass filter, in which high-frequency is suppressed, but also some useful information is lost. Based on the above considerations, both DoG and DoB are performed on image, then $I_{DoG}$ and $I_{DoB}$ are obtained.

### 3.3. Image Fusion

Image fusion is to combine two or more images of the same object to a new image. The new image can describe the object accurately. DoG and DoB have some similar properties, so we can adopt image fusion technology. Considering the advantages and disadvantages of DoG and DoB, we use coefficient selection rule based on the standard deviation (SD) for the fusion of $I_{DoG}$ and $I_{DoB}$ to make a better effect.

Traditional image fusion strategy is to use the fusion rule of average, which can reduce contrast of fused image. SD describes the discrete degree between the value at each pixel and the average value. The bigger the value of SD is, the higher discrete degree of gray value is, and the more information the image contains. Accordingly, we use the weighted average scheme based on SD to fuse images. SD of an image is calculated by

$$\sigma = \sqrt{\frac{\sum \sum [I(x,y) - mean(I)]^2}{N \times M}} \tag{9}$$

$$mean(I) = \frac{\sum \sum I(x,y)}{N \times M} \tag{10}$$

where $N \times M$ is the size of $I$.

The weight of $I_{DoG}$ is:

$$w_{DoG} = \sigma_{DoG} / (\sigma_{DoG} + \sigma_{DoB}) \tag{11}$$

The weight of $I_{DoB}$ is:

$$w_{DoB} = \sigma_{DoB}/(\sigma_{DoG} + \sigma_{DoB}) \tag{12}$$

where $\sigma_{DoG}$ and $\sigma_{DoB}$ represent the SD of $I_{DoG}$ and $I_{DoB}$ respectively. We can achieve the fusion image by

$$I'' = w_{DoG} \cdot I_{DoG} + w_{DoB} \cdot I_{DoB} \tag{13}$$

The purpose of image fusion is to protect more useful information, and the new image can describe the object more accurately and comprehensively.

### 3.4. Local Contrast Equalization

Local contrast equalization（LCE）. The final stage of chain rescales the image intensities to standardize a robust measure of overall contrast or intensity variation. In case of side-lighting, asymmetry in intensity or cast shadow generally appears, we find that some holistic illumination normalization methods cannot handle the side-lighting well, so we should deal with it locally instead of holistically. If we preprocess the image locally, the bad effect by side-lighting might be greatly reduced. Firstly, dividing the image into $n \times n$ blocks evenly. Then we need to deal with image in local blocks. Using the median of the absolute value of the image to reduce extreme values produced by highlights and small dark regions. The mathematical formula is shown below:

$$I''_{c-i}(x,y) = \frac{I''_{cell-i}(x,y)}{(mean(|I''_{cell-i}(x,y)|^a))^{1/a}} \tag{14}$$

$$I''_{cell-i}(x,y) = \frac{I''_{c-i}(x,y)}{(mean(\min(\tau,|I''_{c-i}(x,y)|))^a)^{1/a}} \tag{15}$$

Here, $\alpha$ is a strongly compressive exponent that reduces the influence of large values, $\tau$ is a threshold used to truncate large values after the first phase of normalization. $I''_{cell-i}$ is the *i-th* block of $I''$. Putting $I''_{cell-i}$ together, we can get image $I'''$. After the above transformations, it can still contain extreme values. To reduce their influence on subsequent stages of processing, we apply a final nonlinear mapping to compress over-large values. We use the hyperbolic tangent in the holistic image, thus limiting I to the range $(\tau,\tau)$. The mathematical formula is shown below:

$$I_r(x,y) = \tau g \frac{\exp(I'''(x,y)/\tau) - \exp(-I'''(x,y)/\tau)}{\exp(I'''(x,y)/\tau) + \exp(-I'''(x,y)/\tau)} \tag{16}$$

Because the value of $I_r(x,y)$ may be positive or negative, data regularization is necessary. Then we should normalize the pixel of the image $I_r$ to the range $[0\ 255]$ as follows:

$$R_{result}(x, y) = \frac{I_r(x, y) - \min(I_r)}{\max(I_r) - \min(I_r)} \times 255 \qquad (17)$$

The comparison is shown in Figure 5. LCE can overcome shadow and halo generated under the condition of side-lighting effectively, which makes the image more stable.

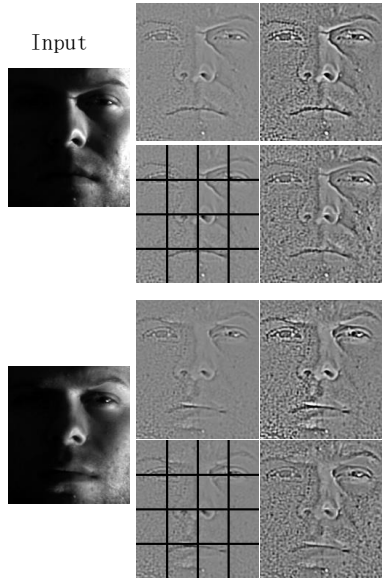

*Fig. 5.* *Comparison of LCE and CE*

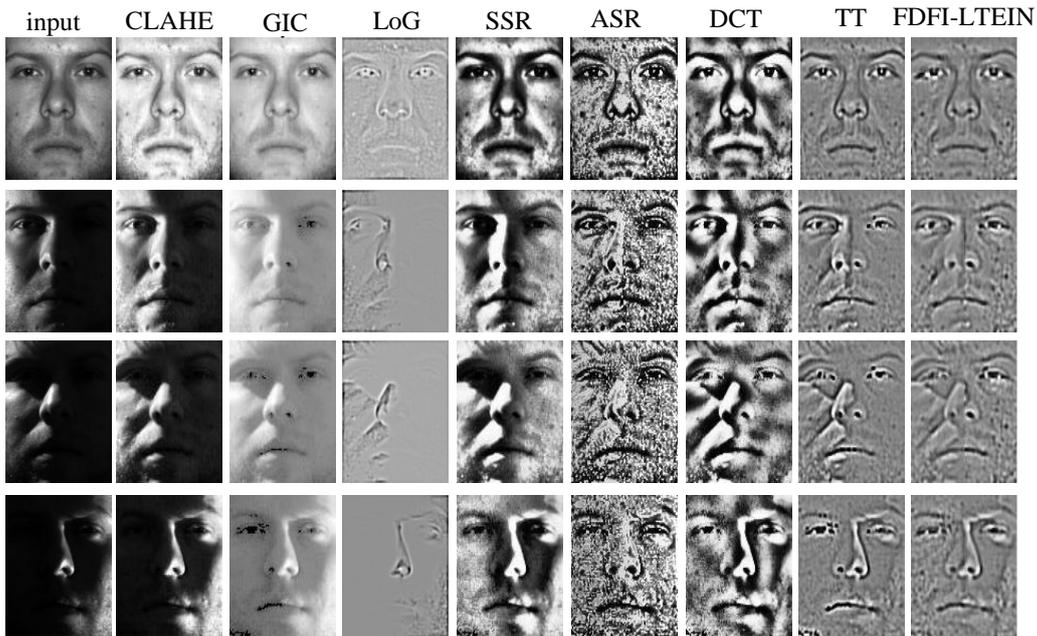

*Fig. 6.* *Comparison with different illumination normalization methods.*

In order to show the advantages of our new method, we select the five images from Extended Yale face database B to compare with other methods, as shown in Fig 6. From left to right: input image, CLAHE, GIC, LoG, SSR, ASR, DCT, TT, FDFI-LTEIN (our preprocessing method). TT and FDFI-LTEIN can reduce the effects caused by the side of light, enhance local texture feature better than others. But halo is generated under the condition of side-lighting by using TT shown in row 3 and 4. The images preprocessed by our methods look more stable under different lighting conditions. In the fourth row, the original image is in a poor lighting condition, the facial features are invisible, and the boundary between shadow region and non-shadow region is enhanced by using FDFI-LTEIN. It can be seen from Fig 6 that the proposed method greatly reduces the influence of lighting variations and enhances local texture feature.

## 4. The proposed Methods

In order to verify the efficiency of our proposed methods techniques, some experiments have been carried out for face recognition by using principal component analysis (PCA), linear discriminate analysis (LDA), local binary patterns histograms (LBPH), multi-scale local uniform binary patterns histograms (MSULBPH). All these methods were applied on the Extend Yale face database B and CMU PIE face database.

*4.1. Data Sets*

The Extended Yale B [30] contains 38 human subjects under 9 poses and 64 lighting conditions. 64 frontal facial images were used to evaluate these illumination compensation methods. The images are divided into five subsets, according to light-source direction: subset 1 (angle < 12° from optical axis), subset 2 (20° < angle < 25°), subset 3 (35° < angle < 50°), subset 4 (60° < angle < 77°) and subset 5 (others). All the images in this database are resized to $100 \times 100$ in our experiment.

The CMU [31] database for pose, illumination and expression (CMU-PIE) consists of data from 68 individuals across 13 different poses, under 43 different illumination conditions, and with 4 different expressions. We focus on the CMU pose subset, which has 43 different illumination images for individuals. And all the faces are the frontal view faces. All the images are resized to $64 \times 64$ in our experiment.

*4.2. Parameter Settings*

In order to select an effective approach to enhance the local dynamic range of the image in dark or shadowed regions while compressing it in bright regions or at highlights, experiments have been conducted on Extended Yale B face database by using local binary patterns histograms (LBPH) and a nearest neighbour classifier with Chi-square as a dissimilarity measure. Results in Table 2 show that $log_2(I+1)$ can achieve a higher recognition rate than others.

**Table 2** Performance of different gray-level enhancement in Extended Yale B

| Enhance Methods | gamma(x,0.2) | In(X+1) | log2(X+1) |
|---|---|---|---|
| LBPH | 0.896 | 0.901 | 0.906 |

In local contrast equalization, we also found that the number of blocks can affect the face recognition rate. The appropriate number of blocks can increase the face recognition rate. But with the number of blocks increasing, some small-scale facial details are lost and side-lighting again appears in the preprocessed face images, the recognition rate is down. Selecting an appropriate number of blocks is very important for face recognition. Some experiments have been conducted on Extended Yale B face database by using LBPH feature and NN classifier as shown in Table 3. To comprehensively consider the recognition rate of test results, we select $n=2$, and the number of blocks is $2^{2\times n}=2^{2\times 2}=16$.

**Table 3** Performance of different gray-level enhancement in Extended Yale B

| The number of blocks | gamma(x,0.2) | In(X+1) | log2(X+1) |
|---|---|---|---|
| LBPH | 0.896 | 0.901 | 0.906 |

*4.3. Experiments Settings*

LBPH has been demonstrated as a powerful local descriptor for microstructures of images. It has certain robustness of light, and widely been used in face recognition. However, the original LBPH operator has same drawbacks in its application to face recognition. It has its small spatial support area, hence the bit wise comparison therein made between two single pixel values is much affected by noise. Moreover, features calculated in the local neighbourhood cannot capture larger scale structure (macrostructure) that may be dominant features of faces [32].

Considering the disadvantages of LBPH, some novel solutions are proposed to improve LBPH. We use multi-scale uniform local binary patterns histograms (MSULBPH) for face recognition to overcome the limitations of LBPH. Firstly, we divide the image preprocessed by illumination normalization methods into $n\times n$ blocks equally. Then, we extract ULBPH features from every block, as shown in Fig 7.

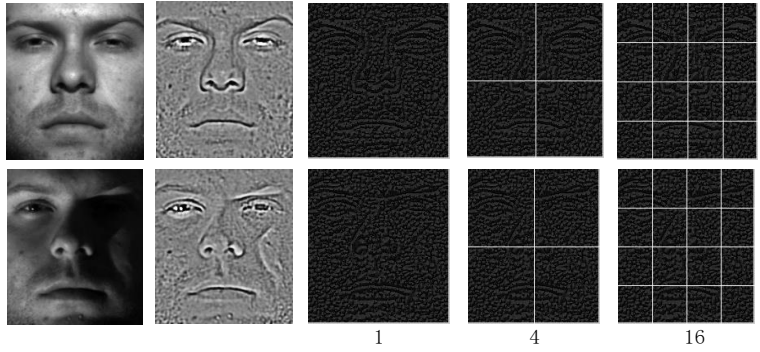

*Fig. 7. Comparison with different illumination normalization methods.*

From Fig 9, we have divided the image into three layers, so we can get 21 blocks. If we divide the image into n layers, the number of blocks:

$$Sum_{block} = \frac{4^n - 1}{3} \qquad (18)$$

We calculate each block of ULBP histograms first, then add all blocks of ULBP histograms to get the face feature, which contains both global and local features.

Thus MSULBPH has several advantages: (1) It is more robust than LBPH in varying illumination conditions for face recognition; (2) it encodes not only microstructures but also macrostructures of image patterns; (3) MSULBPH has a higher dimension than LBPH, and has been demonstrated in [29] that high-dimensional feature and verification performance have a positive correlation.

we compared with 8 different illumination normalization approaches by using PCA, LDA, LBPH and MSULBPH as shown in Tables 4 and 5.

**Table 4** Performance of illumination normalization approaches in Extended Yale B

| Methods | CLAHE | GIC | LoG | MSR | ASR | DCT | TT | Our |
|---|---|---|---|---|---|---|---|---|
| PCA | 0.532 | 0.765 | 0.571 | 0.617 | 0.806 | 0.257 | 0.609 | 0.828 |
| LDA | 0.532 | 0.765 | 0.571 | 0.617 | 0.806 | 0.257 | 0.609 | 0.828 |
| LBPH | 0.532 | 0.765 | 0.571 | 0.617 | 0.806 | 0.257 | 0.609 | 0.828 |
| MS-ULBPH | 0.532 | 0.765 | 0.571 | 0.617 | 0.806 | 0.257 | 0.609 | 0.828 |

**Table 5** Performance of illumination normalization approaches in CMU-PIE

| Methods | CLAHE | GIC | LoG | MSR | ASR | DCT | TT | Our |
|---------|-------|-----|-----|-----|-----|-----|-----|-----|
| PCA | 0.907 | 0.665 | 0.764 | 0.847 | 0.823 | 0.885 | 0.906 | 0.912 |
| LDA | 0.924 | 0.772 | 0.890 | 0.889 | 0.859 | 0.914 | 0.922 | 0.942 |
| LBPH | 0.927 | 0.878 | 0.925 | 0.920 | 0.914 | 0.934 | 0.953 | 0.970 |

Finally, since deep learning has attracted much attention in recent years, we also compared the proposed FDFI-LTEIN with CNN method in [10], and the results are shown in Table 6.

**Table 6** Illumination Normalization Approaches

| Method | Result |
|--------|--------|
| CNN | 0.940[10] |
| Our | 0.948 |

It can be concluded from the above experiments that the performance of FDFI-LTEIN is better than other methods. Therefore FDFI-LTEIN can be used to efficiently enhance the facial texture features and normalize illumination variations for face recognition.

## 5. Conclusion

An enhanced facial texture illumination normalization method is proposed to deal with the critical of illumination variation for face recognition in this paper. By combining DoG and DoB, more useful information is protected. SD rule and LCE strategy are used to reduce halo caused by side-lighting and the boundary between shadow region and non-shadow region. Experimental results also demonstrate that the proposed method is more efficient than the existing illumination normalization methods.

## 6. Acknowledgments

This work was supported in part by the National Natural Science Foundation of China under Grants 61775172, 51805386 and 61371190.The authors acknowledge the anonymous reviewers'insightful and inspirational comments that have greatly helped to improve the technical contents and readability of this paper.